# Quantifying the Ease of Reproducing Training Data in Unconditional Diffusion Models

**Masaya Hasegawa[1], Koji Yasuda[1,2]**

[1]Graduate School of Informatics, Nagoya University
[2]Institute of Materials and Systems for Sustainability, Nagoya University

## Abstract

Diffusion models, which have been advancing rapidly in recent years, may generate samples that closely resemble the training data. This phenomenon, known as memorization, may lead to copyright issues. In this study, we propose a method to quantify the ease of reproducing training data in unconditional diffusion models. The average of a sample population following the Langevin equation in the reverse diffusion process moves according to a first-order ordinary differential equation (ODE). This ODE establishes a 1-to-1 correspondence between images and their noisy counterparts in the latent space. Since the ODE is reversible and the initial noisy images are sampled randomly, the volume of an image's projected area represents the probability of generating those images. We examined the ODE, which projects images to latent space, and succeeded in quantifying the ease of reproducing training data by measuring the volume growth rate in this process. Given the relatively low computational complexity of this method, it allows us to enhance the quality of training data by detecting and modifying the easily memorized training samples.

**Code** — https://github.com/masa-longriver/Quantifying_the_Ease_of_Reproduction

## Introduction

In recent years, diffusion models have led advancements in image generation (Sohl-Dickstein et al. 2015; Ho, Jain, and Abeel 2020), represented by tools like *Stable Diffusion* (Rombach et al. 2022) and *DALL-E* (Ramesh et al. 2022). However, these advancements raise concerns about copyright infringement, as generated images may closely resemble original works by artists.

Diffusion models sometimes generate images nearly identical to their training data (Somepalli et al. 2023a, b). This phenomenon, commonly referred to as memorization, is influenced by the quantity of training data. When the amount of data is limited relative to the model's capacity, models tend to overfit, reproducing specific training samples. This raises a natural question: which images are more easily memorized? Prior studies identified 1,280 memorized images in CIFAR10 (Carlini et al. 2023) and explored the likelihood of generating copyrighted images within training sets (Vyas, Kakade, and Barak 2023). Our study builds on these findings; without such a labor-intensive process, we propose a practical method to quantify the ease of reproducing training data in unconditional diffusion models.

By employing an ordinary differential equation (ODE) to map training samples into the latent space, we establish specific regions corresponding to each sample. When a randomly chosen initial noisy image enters this region, the corresponding training sample is generated. We use the volume growth rate of each training sample along the trajectory to quantify the volume of the region. As this principle relies solely on the diffusion process, the proposed method is applicable to any diffusion models in various domains. The present study not only sheds light on the structure of the latent space, but also has important practical values.

## Related Work

### Score-Based Generative Models

A diffusion model interprets an image as a continuously evolving distribution by progressively adding noise during the forward diffusion process, and the model generates an image by reversing this process (Ho, Jain, and Abeel 2020; Sohl-Dickstein et al. 2015; Song, Meng, and Ermon 2021). Both diffusion and reverse diffusion processes are represented by stochastic differential equations (SDEs) (Song et al. 2021), with forward process in Eq. (1) and its reversed process in Eq. (2).

$$d\boldsymbol{x} = \boldsymbol{f}(\boldsymbol{x}, t)dt + g(t)d\boldsymbol{w} \quad (1)$$

$$d\boldsymbol{x} = -\left[\boldsymbol{f}(\boldsymbol{x},t) - \frac{1}{2}g^2(t)\boldsymbol{s_\theta}(\boldsymbol{x},t)\right]dt$$
$$+\gamma\left[-\frac{1}{2}g^2(t)\boldsymbol{s_\theta}(\boldsymbol{x},t)dt + g(t)d\boldsymbol{w}\right] \quad (2)$$

Here, $\boldsymbol{f}(\boldsymbol{x}, t)$ and $g(t)$ denote the drift and diffusion terms of the SDE, and $\boldsymbol{w}$ represents noise from Brownian motion. The score $\boldsymbol{s_\theta}(\boldsymbol{x}, t)$, representing the time-varying log-gradient of the distribution, allows estimating the data distribution (Alain and Bengio 2014; Hyvärinen 2005; Song and Ermon 2019; Vincent 2011) and is learned by a model like U-Net (Ronneberger, Fischer, and Brox 2015). The parameter

$\gamma$ controls the stochasticity of the reverse process: it is deterministic when $\gamma = 0$, while it is stochastic at $\gamma = 1$.

A commonly used SDE in diffusion models, the Variance Preserving SDE (VPSDE), is defined in Eq. (3),

$$dx = -\frac{1}{2}\beta(t)xdt + \sqrt{\beta(t)}dw \qquad (3)$$

where $f(x, t) = -\beta(t)x/2$ and $g(t) = \sqrt{\beta(t)}$, with $\beta(t) = \beta_{min} + t(\beta_{max} - \beta_{min})$. The perturbation kernel is expressed in Eq. (4).

$$p(x(t)|x(0)) = \mathcal{N}(e^{-m(t)}x(0), v(t)I) \qquad (4)$$

$$m(t) = \frac{1}{2}t\beta_{min} + \frac{1}{4}t^2(\beta_{max} - \beta_{min}) \qquad (5)$$

$$v(t) = 1 - e^{-2m(t)} \qquad (6)$$

## Extract Memorized Training Data

Carlini et al. (2023) proposed a method to identify memorized images within the training data for unconditional diffusion models. They generated a total of $2^{20}$ images from 16 different diffusion models trained on CIFAR10 and calculated the L2 distance between each generated image and every image in the training sets. They found that for memorized images, the generated image is significantly closer to its corresponding training data than to other training samples. The memorization threshold was determined as follows.

$$l(\hat{x}, x; S_{\hat{x}}) = \frac{l_2(\hat{x}, x)}{\alpha \cdot \mathbb{E}_{y \in S_{\hat{x}}}[l_2(\hat{x}, y)]} \qquad (7)$$

Here, $\hat{x}$ denotes the generated image, $x$ represents an image from the training sets being compared, $S_{\hat{x}}$ is the set of $n$ nearest training samples to $\hat{x}$, and $\alpha$ is a hyperparameter. Using this method, they identified 1,280 memorized images from CIFAR10. They also found that the memorized images are almost model independent, implying that it is a peculiarity of the dataset and samples.

# Proposed Method

In this section, we examine the structure of the latent space from the ODE trajectory of each image. Roughly speaking, the latent space is divided into regions for each training sample, and the volume of each region determines the likelihood of image generation. Next, we propose a method to quantify the likelihood by the volume growth rate along the image trajectory.

## The Ease of Reproducing Training Data

Starting from a noisy image drawn from a standard Gaussian distribution, a diffusion model uses the reverse-time SDE in the Langevin equation [Eq. (2)] to generate an image (Song et al. 2021). This stochastic process generates similar but non-identical images for a given initial noise input. Let's focus on the evolution of their average, which is determined by an ODE (Gardiner 1985).

$$dx = \left[f(x, t) - \frac{1}{2}g^2(t)s_\theta(x, t)\right]dt \qquad (8)$$

This is the time-reversal of Eq. (2) with deterministic ($\gamma = 0$) condition. This ODE establishes the 1-to-1 mapping between images and their noise counterparts. Next, consider choosing a (hyper)sphere region $S$ of small radius $\sigma$ centered on one of the training images. The ODE transforms each point in $S$ to a noisy image, causing volume changes over time. On one hand, recognizable images occupy a small volume in the high-dimensional space. On the other hand, the diffusion model transforms a randomly drawn noise into a recognizable image. Thus, we hypothesize that recognizable images expand faster than others under the ODE.

Using $m$ in Eq. (4) in place of time $t$, the diffusion process can be expressed compactly as

$$dx = -[x + s(x)]dm \qquad (9)$$

$$s(x) = \nabla \log P(x) \qquad (10)$$

$$P(x) = \frac{1}{Z}\sum_i \exp\left[-\frac{\left(x - y^{(i)}e^{-m}\right)^2}{2v}\right] \qquad (11)$$

Here, $y^{(i)}$ represents the $i$-th training sample, $P(x)$ is the distribution of the noised one, and $Z$ is the normalization constant.

Eqs. (9)-(11) reveal interesting properties of the diffusion process. First, if there is only one training sample $y$, the exact solution for $x(m)$ is:

$$x(m) = ye^{-m} + \sqrt{v}C \qquad (12)$$

where $C$ is an arbitrary vector that we take as a point on a sphere. The data point $y$ moves toward the origin as $e^{-m}y$, and the distance between $x(m)$ and $y(m) = e^{-m}y$ increases monotonically as $\sqrt{v}$, except when initial $x(0^+)$ is exactly $y(0^+)$. In short, $x(m)$ is repelled from $y(m)$.

When there are two training samples, each sphere around it moves approximately along with its center $e^{-m}y^{(i)}$ and expands as $\sqrt{v}$. However, the sphere $S^{(1)}$ is also repelled by $y^{(2)}(m)$, resulting in distortion. Now consider a point $x$ on the line segment between endpoints $e^{-m}y^{(1)}$ and $e^{-m}y^{(2)}$. At the midpoint, the scores cancel out. Hence, the right half of the line segment always stays in the right half.

Unfortunately, analyzing the general case with more data points is challenging by hand. However, we hypothesize that each initial sphere $S^{(i)}$ will project to a distinct, non-overlapping region $R^{(i)}$ in the latent space. Figure 1 shows an example where each training data point diffuses into distinct, non-overlapping regions in the latent space under the exact score. If an initial noisy image is chosen from $R^{(i)}$, it will reverse diffuse into the neighborhood of $i$-th image, meaning that the volume of $R^{(i)}$ represents the probability of generating that image. Thus, analyzing the memorization in a diffusion model deduces calculating the volume growth rate along the ODE trajectory.

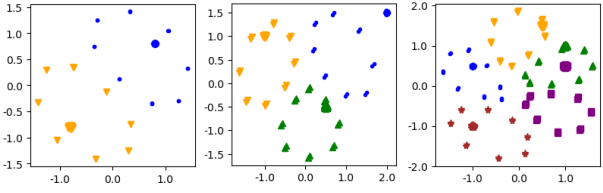

Figure 1: Time evolution of samples in two-dimensions by Eq. (9) under exact score. The number of samples is 2, 3, and 5 from left to right. The small (large) circles represent samples at time $t = 10^{-3}$ ($t = 1$).

## Quantifying the Ease of Reproduction

The volume change of a system following an ODE has long been studied in dynamical systems in relation to chaos. In our equation, the volume change is determined by the divergence of the score, and recent studies reported a method to approximate such higher-order derivatives (Meng et al. 2021). We examined various models for the second derivatives but faced significant challenges. Therefore, inspired by the concept of the Lyapunov exponent (Greiner 2009), we propose a method to quantify the ease of image reproduction, as presented in Algorithm 1.

Our approach begins with a central data point and placing surrounding points orthogonally on a small sphere around it. We then evolve these points for a short period according to the ODE, and the volume of the resulting parallelepiped provides the volume growth rate. The stretch rate along each axis at each diffusion step, reflecting how far the surrounding points move from the center, determines the volume growth rate. The cumulative product across steps yields the total volume growth rate, which can be formulated as follows:

$$l_t = \prod_t \prod_k \frac{\|x_{t+1}^{(k)} - x_{t+1}^{(0)}\|}{\|x_t^{(k)} - x_t^{(0)}\|} \qquad (13)$$

where $l_t$ denotes the volume growth rate from time $0$ to $t$, $x_t^{(0)}$ is the noisy ground-truth image at time $t$, and $x_t^{(k)}$ represents the $k$-th surrounding point in each axis direction. We orthogonalize $N$ surrounding points $x_0^{(1:N)}$ on a spherical radius $\sigma$ and diffuse $T$ steps. Due to the high dimensionality of images, we compute this rate on a logarithmic scale for numerical stability. At each diffusion step, we employ Gram-Schmidt orthogonalization (Hogben 2013), aligning with the axis of largest growth and resetting the length to its initial value. While we ideally use the number of surrounding points $N$ equal to the image dimensions to compute the full volume growth rate, we have found that fewer points suffice if relative growth rates are needed, as shown in the next section.

---

**Algorithm 1:** Calculation of log volume expand rate

**Require**:
  $x_0^{(0)} \in \mathbb{R}^D$: Target data (D: Dimension)
  T: The number of diffusion steps.
  N: The number of axes.
  $\sigma$: Small sphere size of surroundings.

1: let $L = [\log l_1, \ldots, \log l_T]$
2: let $X_t = [x_t^{(1)}, \ldots, x_t^{(N)}]$
3: $\epsilon \sim \mathcal{N}(\mathbf{0}, I) \in \mathbb{R}^{N \times D}$
4: $X_0 \leftarrow x_0^{(0)} + GramSchmidt(\epsilon) * \sigma$
5: **for** $t = 0$ **to** $T - 1$ **do**
6:     $x_{t+1}^{(0)} \leftarrow forwardODE(x_t^{(0)})$ # eq. (8)
7:     $X_{t+1} \leftarrow forwardODE(X_t)$ # eq. (8)
8:     $\log l_{t+1} \leftarrow \log l_t$
9:     **for** $k = 1$ **to** $N$ **do**
10:       $\log l_{t+1} \leftarrow \log l_{t+1} + \log \|x_{t+1}^{(k)} - x_{t+1}^{(0)}\|$
             $- \log \|x_t^{(k)} - x_t^{(0)}\|$
11:     Sort $X_{t+1}$ in descending order based on
         $\|X_{t+1} - x_{t+1}^{(0)}\|$
12:     $X_{t+1} \leftarrow GramSchmidt(X_{t+1} - x_{t+1}^{(0)})$
13:     $X_{t+1} \leftarrow X_{t+1} * \sigma + x_{t+1}^{(0)}$
14: **return** $L$

---

## Experiments

In this section, we first validate our proposed method by calculating the volume growth rate of training data in an overfitted model. We compare images included in the training set (images that can be generated) with those not included (images that cannot be generated). In another experiment, we analyze the volume growth rates of memorized images identified in previous study, comparing them to unmemorized images and demonstrating that memorized images show a higher likelihood of being generated. Finally, we investigate the behavior of our proposed method under various parameter settings to assess its robustness and explore more efficient approaches for measuring the ease of reproduction.

### The Ease of Reproduction in an Overfitted Model

When the amount of training data is limited, diffusion models overfit, generating only the images included in training data (Zhang et al. 2024). To validate our method, we used an overfitted model trained on a small dataset to compare the volume growth rates of trained and non-trained images. We randomly selected $2^6$ CIFAR10 images, applied horizontal flipping to create $2^7$ images for training, and set the number of epochs to 300,000 to induce overfitting. Other settings are the same as Song et al. (2021). Volume growth rates were calculated for $2^7$ trained and non-trained images

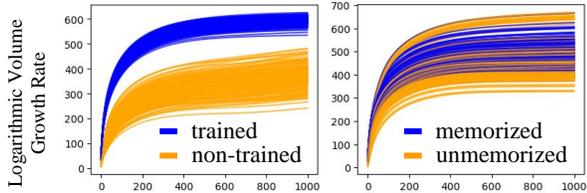

Figure 2: Results of the volume growth rate. (Left) rate in log scale for training and non-training images diffused with an overfitted model. (Right) same for memorized and unmemorized images.

with parameters $T = 1000$, $\sigma = 0.05$, and $N = 100$. The results are shown in Figure 2.

As seen in the figure, there is a clear difference in volume growth rates between trained and non-trained images. The trained images exhibit an average volume growth rate of $e^{594}$, whereas the untrained images show only $e^{373}$. Thus, the proposed method is a valid metric for quantifying the ease of image reproduction.

### The Ease of Reproducing Memorized Images

Here, we compare the volume growth rates of 1,280 memorized and unmemorized CIFAR10 images (as identified by Carlini et al., 2023) to discuss the reproductive ease of memorized images. The 50,000 CIFAR10 images were horizontally flipped and doubled to 100,000. We trained the model on them for 1,500 epochs, keeping other settings the same as before, *which wasn't overfitted*. Volume growth rate for 1,280 memorized and unmemorized images were measured with parameter $T = 1000$, $\sigma = 0.05$, and $N = 100$. The results are shown in Figure 2.

Memorized images generally exhibited higher volume growth rates. A t-test confirmed significant differences between memorized and unmemorized images at the 1% level, indicating that memorized images are easier to reproduce. Therefore, once measuring the volume growth rates of training data, we can calculate the ease of reproduction by its comparison.

Additionally, some unmemorized images also showed high volume growth rates. These images often have distinctive characteristics, such as being monochromatic, having simple backgrounds with small objects. Examples of these images are provided in Figure 6. Our method easily identifies such images, but we can't find the reason for their high rates. We leave it our future work.

### Investigating Various Parameter Settings

Finally, we investigate the behavior of the proposed method under different parameter settings, demonstrating its robustness and exploring more efficient ways to quantify the ease of reproduction.

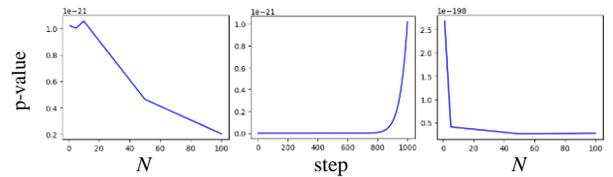

Figure 3: p-values when varying the number of axes $N$ and steps in experiments comparing the volume growth rate of memorized and unmemorized images. (Left) p-values for each $N$ at step=1000. (Center) p-values for each step at $N$=1. (Right) p-values for each $N$ at step=1.

Results on varying the number of axes $N$ and sphere size $\sigma$ are in Appendix A. The shape of the plot remained consistent across values of $N$, suggesting that reproductive ease can be quantified without full-dimensional calculations. Additionally, varying $\sigma$ between 0.001 and 0.1 did not impact the ease of reproduction.

We then compared the t-test p-values for different axis counts $N$ and diffusion steps $t$ in calculating the volume growth rate (Figure 3). While higher $N$ provides more precise quantification, even a single axis yields sufficiently small p-values. Furthermore, using fewer steps results in smaller p-values. These findings indicate that quantifying the ease of reproduction requires *only the stretch rate along a single axis over one step*, which has practical applications. Using this approach, we selected the top 1,280 images from CIFAR10, categorizing as either easy-to-memorize or hard-to-memorize, as shown in Figures 7 and 8. The results reveal that the easy-to-memorize images tend to have simple compositions, while the hard-to-memorize images are more complex.

### Conclusion

In this study, we proposed a method to quantify the ease of reproducing training data. Images are projected into a specific region of the latent space via the ODE, and the regions they occupy determine generation probability. So, we proposed calculating the volume growth rate to quantify the ease of reproduction. We validated our method through two experiments, trained images in overfitted model and memorized images of previous work. Our parameter experiments demonstrated that only the stretch rate along a single axis over one step is sufficient to measure the ease of reproduction. This straightforward and effective method enables us to find easy reproduced images in training sets. Although we have demonstrated the method in the image domain, it is applicable to any diffusion models based on SDEs. Future work includes investigating the behavior of this method in stochastic diffusion process, latent diffusion models, conditional diffusion models, and the diffusion models in other domains.

# Appendix

## A. Experimental Results Across Various Parameter Settings

We present the results of the growth rate with varying the number of axes $N$ and the small sphere size $\sigma$. We tested four varies of $N = \{1, 10, 50, 100\}$, and four varies of $\sigma = \{0.001, 0.01, 0.05, 0.1\}$. Figure 4 shows the results for the trained and non-trained images in the overfitted model used in the first experiment. Figure 5 presents the results for the memorized and un-memorized images from the second experiment. The overall shape of the plot remains consistent across all combinations of $N$ and $\sigma$, indicating that this method is robust to changes in both parameters, $N$ and $\sigma$.

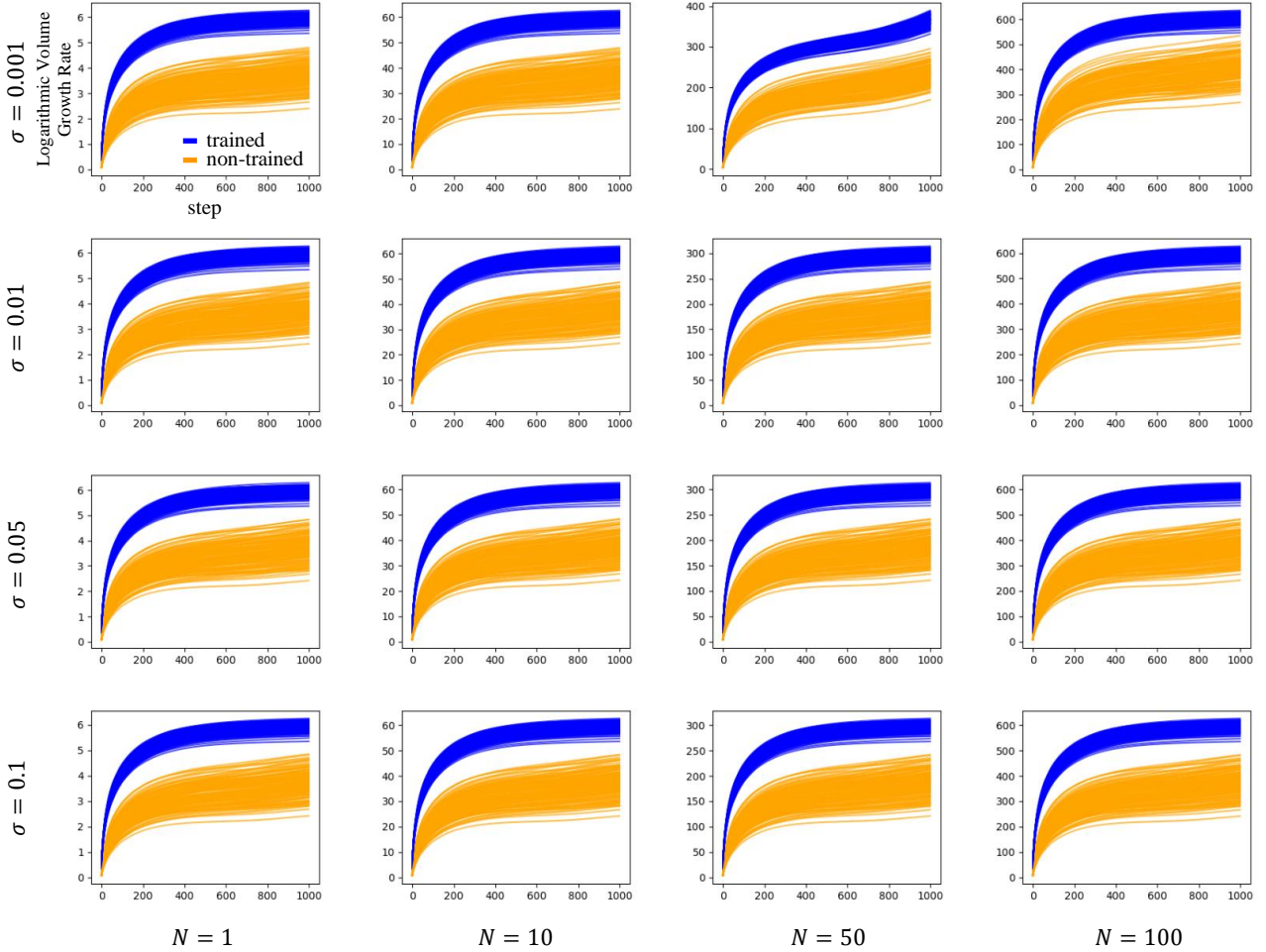

Figure 4: Volume growth rates of trained and non-trained images when varying the number of axes $N$ and the small sphere size $\sigma$.

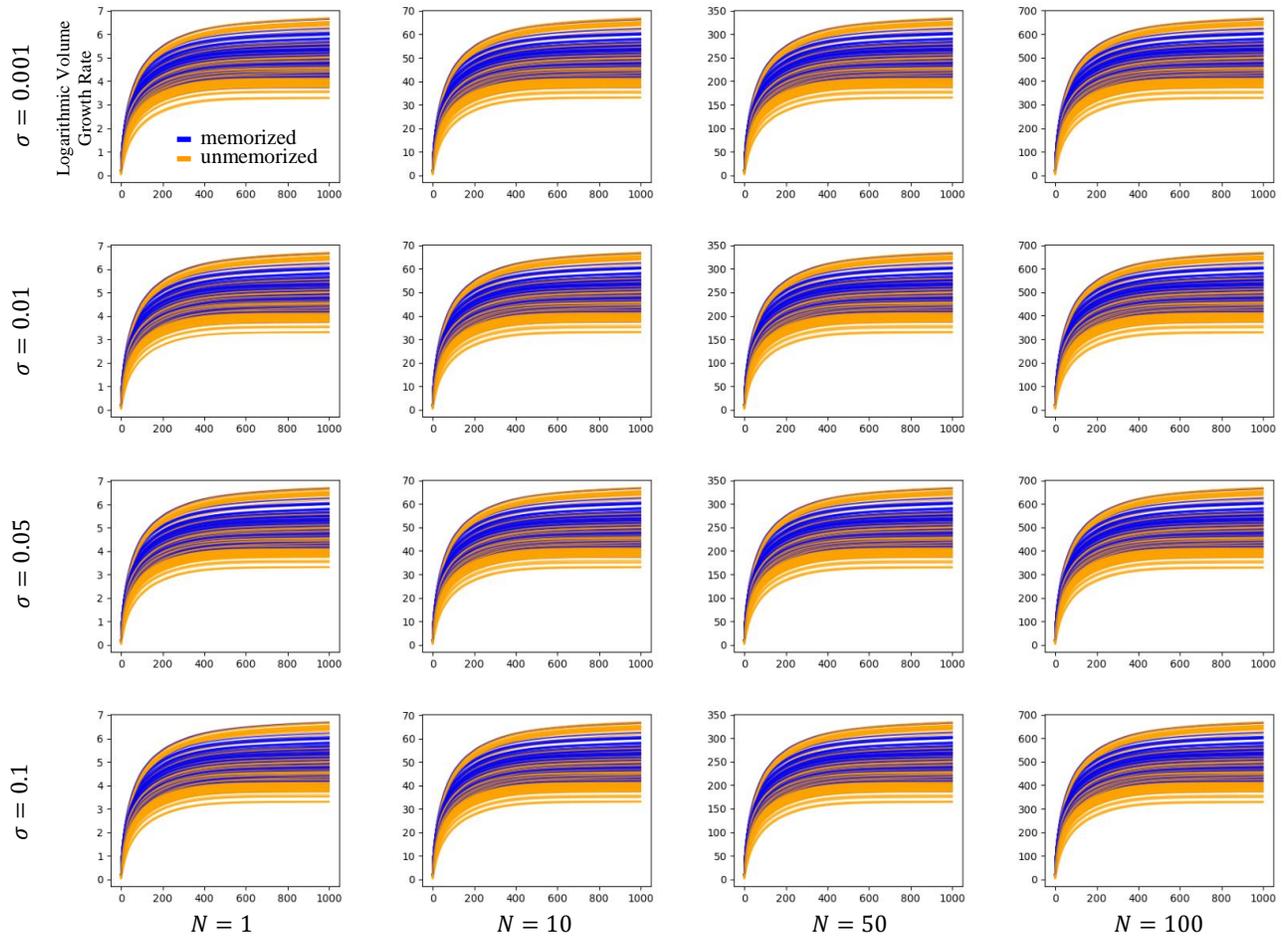

Figure 5: Volume growth rates of memorized and unmemorized images when varying the number of axes $N$ and the small sphere size $\sigma$.

## B. Sample Images Assessed for Ease of Generation

### B-1. Unmemorized Images with Large Volume Expand Ratio

In the second experiment comparing the volume growth rates of memorized and unmemorized images, we found some unmemorized images recorded high volume growth rates. These are shown in Figure 6. These images have common characteristics such as being monochromatic or containing small objects with large background areas.

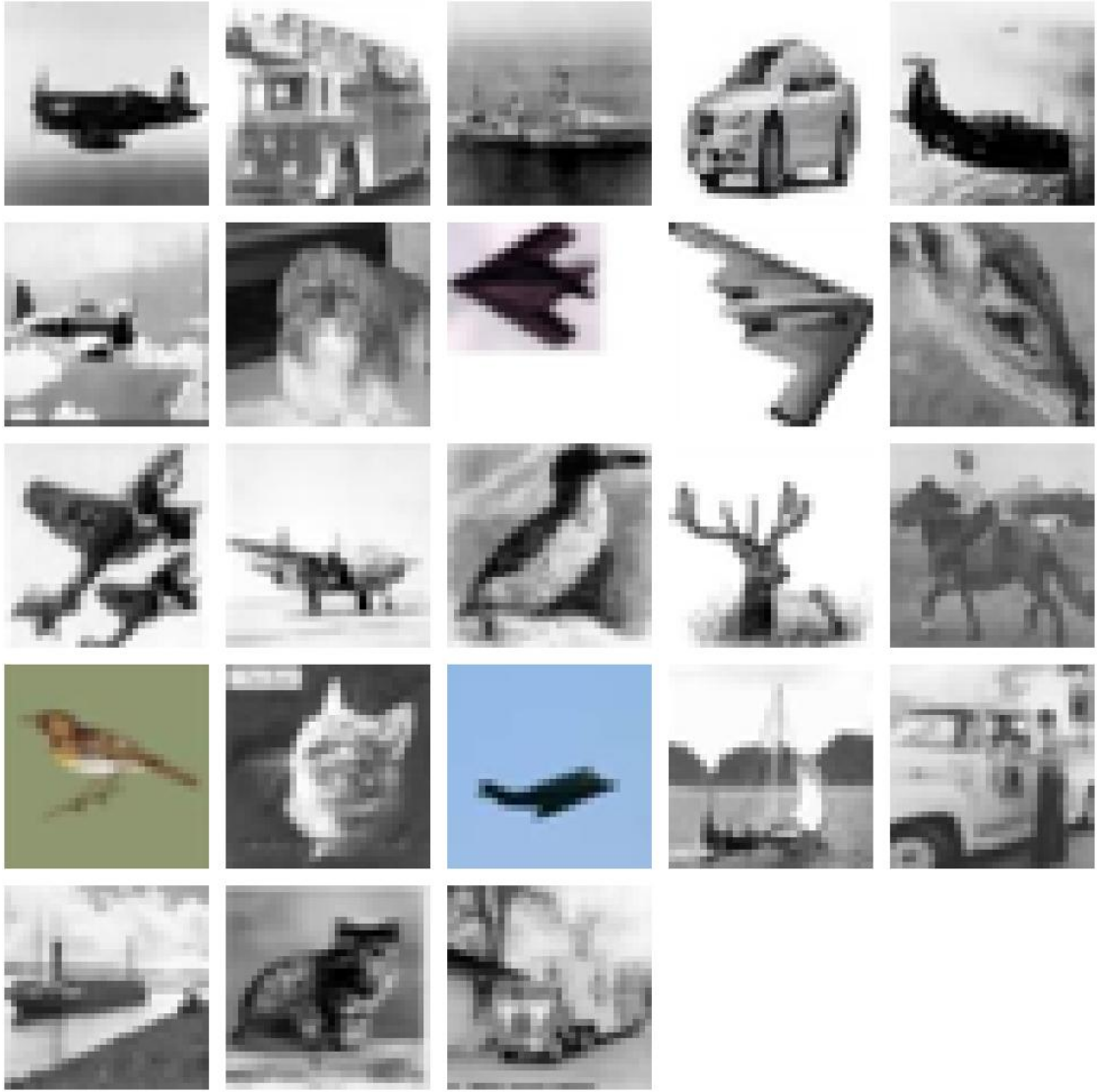

Figure 6: Unmemorized images with high volume growth rate. From generated images with $N = 100$, $\sigma = 0.05$, and $step = 1000$, we select those with growth rate greater than $e^{550}$.

### B-2. Example of Easily and Hardly Generated Images in CIFAR10

Examples of easily and hardly generated images in CIFAR10, as identified by our method, are shown in Figures 7 and 8. We augmented the 50,000 images to 100,000 by horizontal flipping to train a diffusion model. Then we calculated their volume growth rates with $N = 1$, $\sigma = 0.05$, and $step = 1$, selecting the top and bottom 1,280 images.

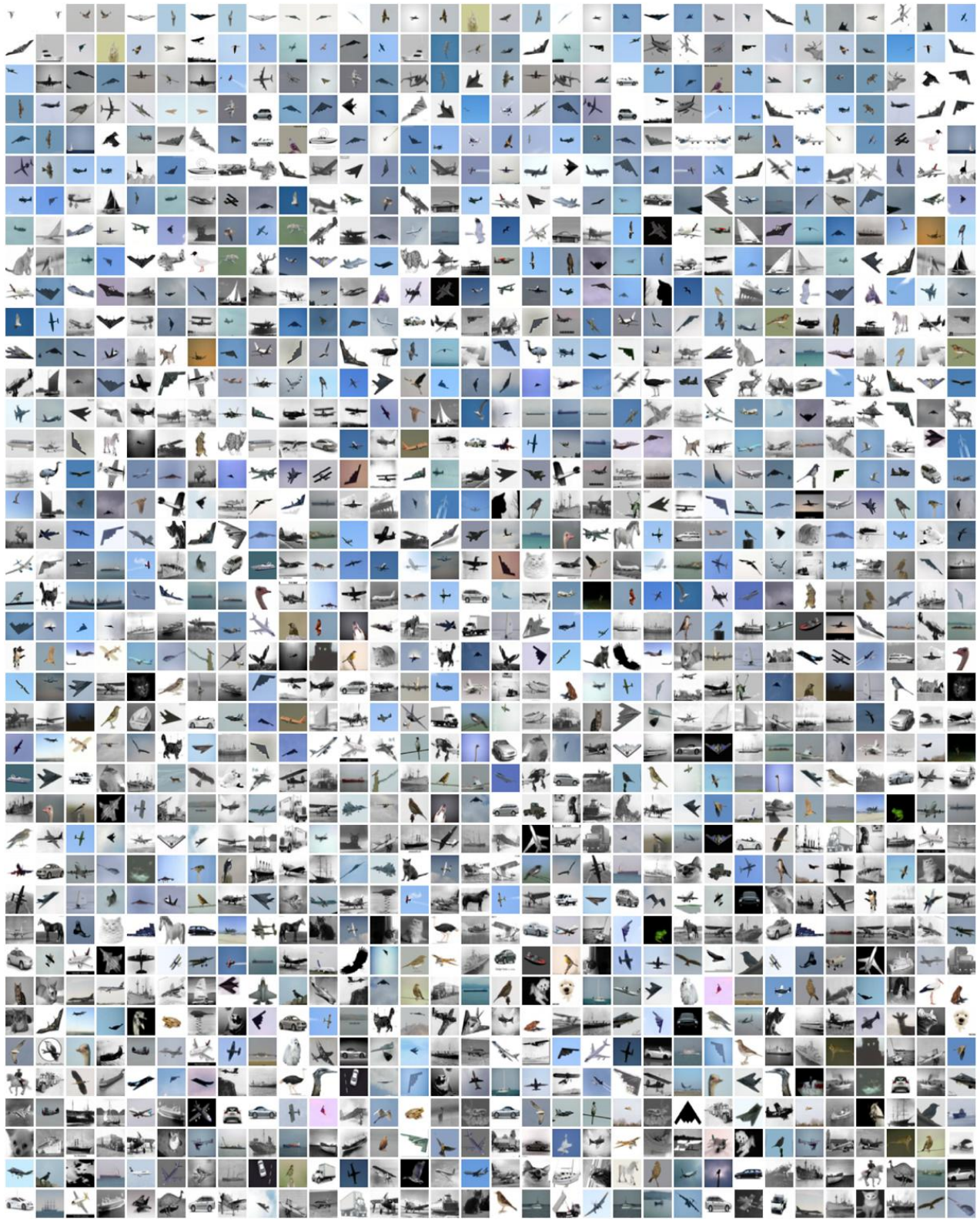

Figure 7: Easily reproduced images extracted from our proposed method. We calculate volume growth rates with $N = 1$, $\sigma = 0.05$, and $step = 1$, then select the top 1,280 images.

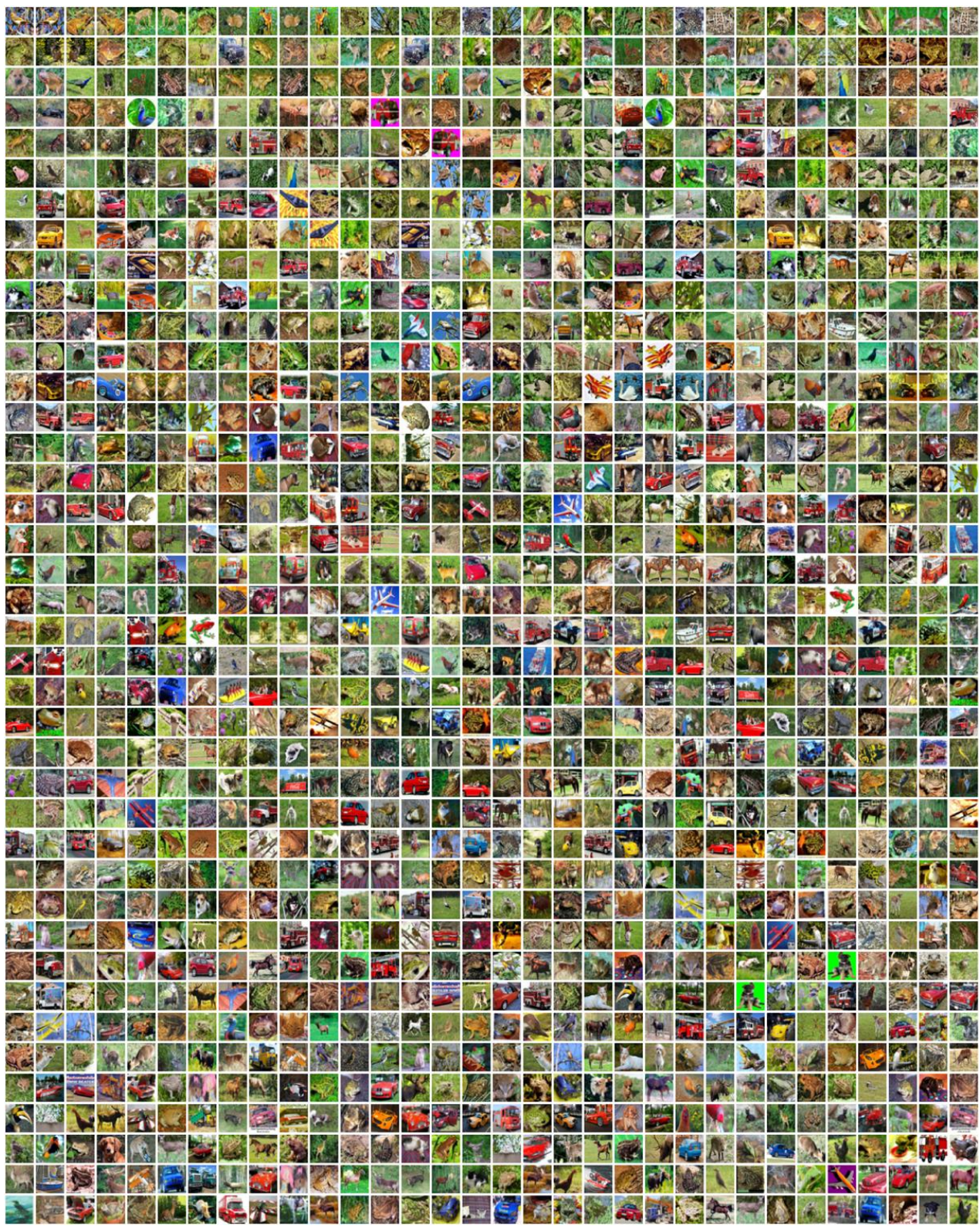

Figure 8: Hardly reproduced images extracted from our proposed method. We calculate volume growth rates with $N = 1$, $\sigma = 0.05$, and $step = 1$, then select the bottom 1,280 images.